\documentclass[11pt,a4paper]{article}
\usepackage[hyperref]{mwpsurvey}
\usepackage{times}
\usepackage{latexsym}
\usepackage{graphicx}
\usepackage{amsmath}

\usepackage{microtype}

\aclfinalcopy % Uncomment this line for the final submission
\title{
Towards Tractable Mathematical Reasoning: Challenges, Strategies, and Opportunities
}

\title{
Towards Tractable Mathematical Reasoning: Challenges, Strategies, and Opportunities for Solving Math Word Problems
}

\author{Keyur Faldu \thanks{\hspace{0.1cm}Correspondence to k@embibe.com}\\
  Embibe \\\And
  Amit Sheth \\
  University of\\ South Carolina \\\And
  Prashant Kikani \\
  Embibe \\\And
  Manas Gaur \\
  University of\\ South Carolina \\\And
  Aditi Avasthi \\
  Embibe \\
  }

\date{}

\begin{document}
\maketitle
\begin{abstract}

\end{abstract}

Mathematical reasoning would be one of the next frontiers for artificial intelligence to make significant progress. The ongoing surge to solve math word problems (MWPs) and hence achieve better mathematical reasoning ability would continue to be a key line of research in the coming time. We inspect non-neural and neural methods to solve math word problems narrated in a natural language. We also highlight the ability of these methods to be generalizable, mathematically reasonable, interpretable, and explainable. Neural approaches dominate the current state of the art, and we survey them highlighting three strategies to MWP solving: (1) direct answer generation, (2) expression tree generation for inferring answers, and (3) template retrieval for answer computation. Moreover, we discuss technological approaches, review the evolution of intuitive design choices to solve MWPs and examine them for mathematical reasoning ability. We finally identify several gaps that warrant the need for external knowledge and knowledge-infused learning, among several other opportunities in solving MWPs.

\begin{figure}[h]
    \centering
    \includegraphics[scale=0.32]{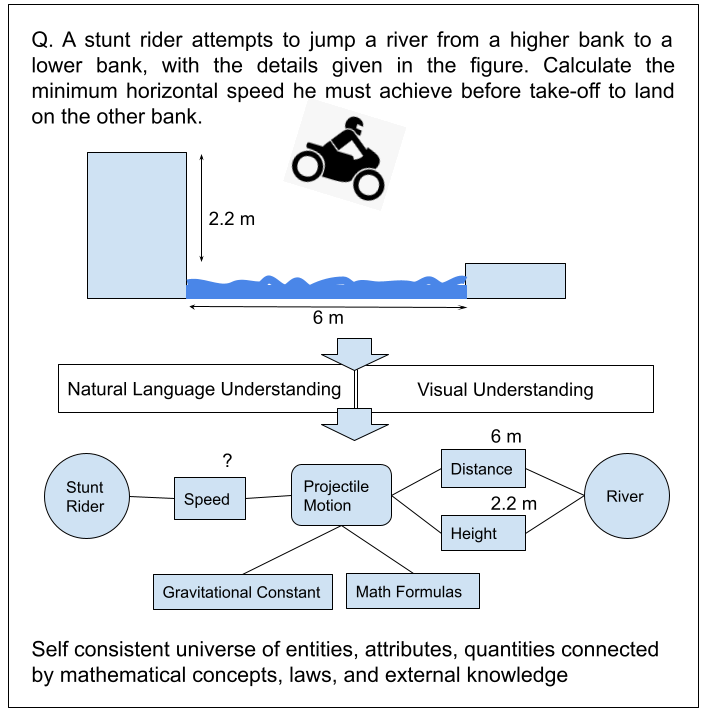}
    \caption{An illustration of the mathematical reasoning process that involves understanding of natural language and visual narratives. Essentially, the process comprises extracting entities with explicit and implicit quantities building a self-consistent symbolic representation invoking mathematical laws, axioms, and symbolic rules. 
}
    \label{fig:math_reason}
\end{figure}

\section{Introduction}

The recent success of artificial intelligence over a wide range of natural language processing tasks like hypothesis entailment, question answering, summarization, translation, etc., has posed the question that \textit{how intelligent are these systems?}. The General Language Understanding Evaluation (GLUE) \cite{wang2018glue}, and SuperGLUE \cite{wang2019superglue} benchmarks have been created to assess their natural language understanding abilities. Soon enough, deep learning systems have outperformed human-level performance over these benchmarks.  However it is not understood if such performance is attributed to underlying reasoning capabilities \cite{sap2020commonsense}\cite{storks2019commonsense}\cite{khashabi2019capabilities}. Discovering and learning salient heuristic patterns to produce a likely outcome is also much better than rule-based systems. When learning is achieved without reasoning, it will reduce to mimicking, which would not be generalizable \cite{bandura2008observational}. Reasoning ability is at the center of performing higher-order complex tasks. Mathematical reasoning is one of the niche abilities humans possess \cite{lithner2000mathematical} \cite{english2013mathematical}, it constitutes linguistic reasoning, visual reasoning, common sense reasoning, logical reasoning, numerical reasoning, and symbolic reasoning. Humans do not learn mathematical reasoning just based on experience and evidence, but it results from learning, inferring, and applying laws, axioms, and symbolic rules. Mathematical reasoning could be one of the most important benchmarks to test the progress made by artificial intelligence. 

\begin{figure*}[h]
    \centering
    \includegraphics[scale=0.31]{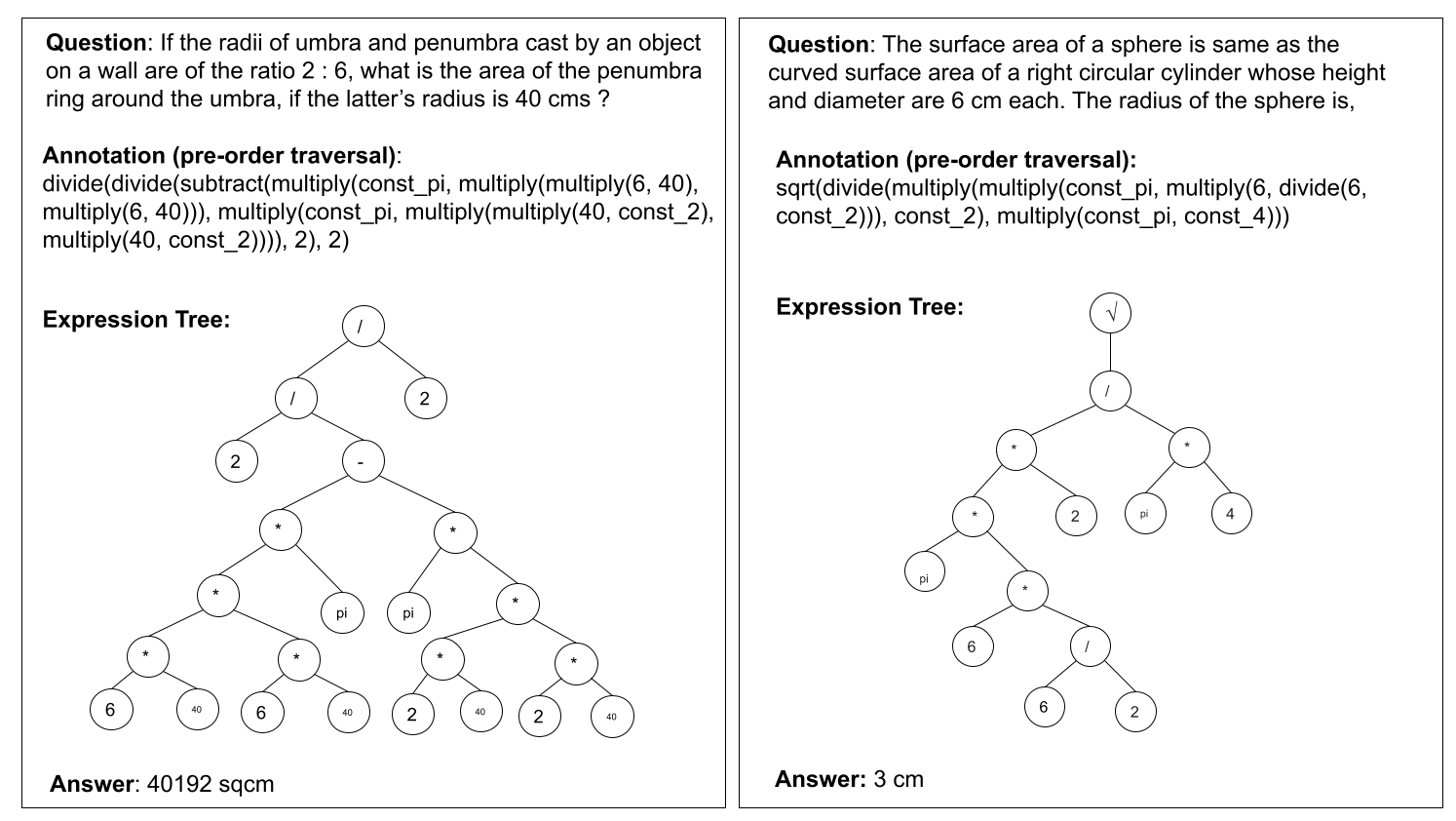}
    \caption{Example of relatively complex MWPs from dataset MathQA \cite{amini2019mathqa}, requiring understanding of specific words like umbra and penumbra, applying multiple math concepts, and accessing mathematical knowledge like math formula for area of circle, sphere, value of $\pi.$}
    \label{fig:complexmwp}
\end{figure*}

\textit{Mathematical reasoning ability involves understanding natural language and visual narratives, extracting explicit and implicit quantities using common sense and domain-specific knowledge, inferring a self-consistent symbolic universe, and exploiting mathematical laws, axioms, and symbolic rules} \cite{amini2019mathqa}\cite{saxton2019analysing}. The problems requiring mathematical reasoning ability pose a unique opportunity to build a system that requires collaborative interactions between neural systems and symbolic systems \cite{kahneman2011thinking}. AI models for natural languages understanding and visual understanding could be developed using fast or probabilistic systems, and corresponding mathematical laws, axioms, and rules could be executed using slow and deterministic systems. Math word problems (MWP) are natural language questions that require mathematical reasoning to solve them. \textit{MWP is the natural language narrative that contains a self-consistent universe where entities with quantitative information interact with each other under well-defined mathematical axioms, laws, and theorems. Answering MWP would require solving unknown quantities or sets of unknown quantities from known explicit or implicit quantities using semantic information, commonsense world knowledge, and mathematical domain knowledge.} Thus, MWPs can be of varying difficulty levels based on the complexity of their narratives, implicit world and domain knowledge needed to solve the problem, and the application of mathematical topics and their interactions. Figure \ref{fig:complexmwp} demonstrates couple of examples of relatively complex MWPs.

The ability to solve MWPs in an automated fashion could lead to many applications in the education domain for content creation and delivering learning outcomes \cite{donda2020framework} \cite{faldu2020adaptive}. For example, it could aid students’ learning journey by resolving their doubts on the fly; potentially solve the problem of content deficiency for practice assessments \cite{dhavala2020auto}. Particularly by complementing MWPs recommendation systems with AI-based solution generation system \cite{faldu2020system}.

The task of solving MWPs has a long history with consistent research attention. This article surveys the progress in solving MWPs, including datasets, approaches, and architectures, broadly divided into non-neural and neural techniques. We also analyze the mathematical reasoning ability of such methods.

\begin{figure}[h]
    \centering
    \includegraphics[scale=0.37]{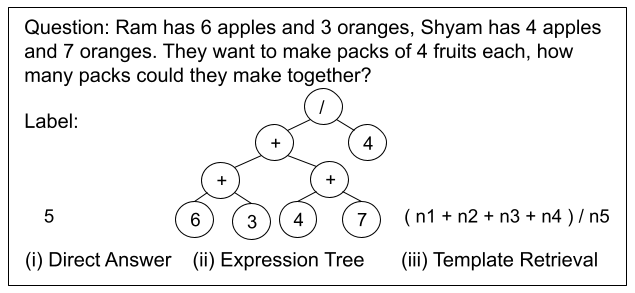}
    \caption{Example of math word problem and its possible problem formulations}
    \label{fig:mwp}
\end{figure}

\section{Background and Non-Neural Approaches}

The complexity of solving MWPs has been well understood for a long time, demanding logical systems representation and machine’s ability to apply such systems with natural language understanding and generative capabilities \cite{feigenbaum1963computers}. Before the surge of neural network-based approaches to solving MWPs, several non-neural techniques were explored since the 1960s, classified into rule-based or pattern matching, semantic parsing, statistical, and machine learning approaches. However, most non-neural methods focus on the template retrieval approach or predicting the answer using rules. These approaches are explained in the following sections.

\subsection{Rule-based or Pattern Matching Approaches}
Normalizing and transforming input narratives using heuristics, such as rules, regular expressions, or pattern matching, was the first attempt in solving MWPs \cite{feigenbaum1963computers} \cite{bobrow1964natural} \cite{charniak1968calculus}. Such approaches are very intuitive but could only handle limited scenarios, and hence, they lack the flexibility in handling simple variations of math word problems. Further, efforts have been made to develop schema, frame, or extract entities and quantities associated with it to apply rules deterministically. \cite{fletcher1985understanding} \cite{dellarosa1986computer} \cite{yuhui2010frame}. However, such rule-based and pattern matching algorithms would not be scalable to learn general-purpose mathematical reasoning ability.

\subsection{Semantic Parsing}

A semantic parsing-based system attempts to learn the semantic structure of the input problem and transforms it into mathematical expressions or a set of equations. It represents problems in object-oriented structure, much similar to text-SQL generation, following methods in natural language processing \cite{liguda2012modeling} \cite{koncel2015parsing}. Underlying tree dependencies of mathematical semantics are captured by context-free grammar rules, which is analogous to dependency parse trees from the perspective of mathematical semantics \cite{shi2015automatically}.

\subsection{Statistical and Machine Learning Approaches}

Prior works on Statistical and machine learning approaches developed automated methods that are rule-based and semantic parsing, which processes training data to derive a set of templates or learn the semantic structure from the input. They range from text classification of MWPs with equation templates, extracting and mapping quantities with equations, or entities extraction and classifying their relationships \cite{kushman2014learning}.  These methods use machine learning algorithms like support vector machines, probabilistic models, or margin classifiers for concept prediction, equation prediction, or slot filling. They leverage sentence semantics or verb categorization to map mathematical operations with segments of an MWP \cite{hosseini2014learning} \cite{amnueypornsakul2014machine}. A further set of candidate hypotheses are reduced by handling noun slots and number slots separately \cite{zhou2015learn}. Few other techniques process narratives to fill concepts-specific slots first and derive equations from them using domain-specific rules \cite{mitra2016learning}\cite{roy2018mapping}.

\section{Mathematical Reasoning in Non-Neural Approaches}

Mathematical reasoning ability is the key to solving mathematical problems. There were few attempts made in this direction using reasoning-oriented sub-tasks like quantity entailment for semantic parsing \cite{roy2015reasoning}, comprehending numerical information, quantity alignment prediction for slot filling , handling quantity slots and noun slots separately, ignoring extraneous information present in narratives, etc \cite{bakman2007robust} \cite{mitra2016learning} \cite{roy2018mapping} \cite{zhou2015learn}.

Non-neural methods that consider rule-based, pattern matching, semantic parsing, and machine learning-based approaches are limited to specific areas of mathematics like addition-subtractions, arithmetic, linear algebra, calculus, or quadratic equations. The performance of such methods is constrained by the coverage and diversity of MWPs in the training corpus. The performance of such approaches could not scale on out-of-corpus MWPs mainly because of the relatively small-sized training corpus \cite{koncel2016mawps} \cite{kushman2014learning}. \cite{huang2016well} attempted these techniques on a large-scale dataset Dolphin18K, having 18000 annotated linear and non-linear MWPs. Non-neural methods performed much worse on diverse and large-scale datasets than their reported performance on their corresponding small and specific datasets, and their performance improved sub-linearly with more extensive training data \cite{huang2016well}. It made a solid case to build more generic systems with better reasoning ability. 

\begin{table*}[hbt!]
\fontsize{10pt}{15pt}
\selectfont
\begin{tabular}{p{0.58\linewidth} | p{0.08\linewidth} |  p{0.24\linewidth}}
\hline Addition-Subtraction Math Word Problems & Subtype & Semantic Slots \\
\hline
Sam’s dog had 8 puppies. He gave 2 to his friends. He now has 6 puppies . How many puppies did he have to start with? & Change & (i) start
(ii) end 
(iii) gains
(iv) losses \\
\hline
Tom went to 4 hockey games this year, but missed 7 . He went to 9 games last year.  How many hockey games did Tom go to in all ? &
Part-Whole & 
(i) whole
(ii) parts \\
\hline
Bill has 9 marbles. Jim has 7 more marbles than Bill. How many marbles does Jim have? & Compare & (i) large quantity
(ii) small quantity
(iii) difference \\
\hline
\end{tabular}
\caption{Classifying math word problems into subtypes, and semantic slots required for solving MWPs in each subtypes.
}
\label{tab:expdomain}
\end{table*}

\subsection{Domain Knowledge or External Knowledge}

Classifying MWPs into subtypes and handling them using mathematical domain knowledge is another way to further push such systems' capabilities. It becomes easier to reason about a problem classified to a subtype as it reduces the candidate math laws, axioms, and symbolic rules and contextualizes semantics related to the specific area of math \cite{mitra2016learning} \cite{roy2018mapping}\cite{bakman2007robust} \cite{amnueypornsakul2014machine}. For example, addition and subtraction problems can be grouped in three classes of mathematical concepts, (i) change, (ii) part-whole and (iii) compare as shown in table \ref{tab:expdomain} \cite{mitra2016learning}. Semantic parsing techniques form rules to extract semantic information specific to these sub-types. Table \ref{tab:expdomain} illustrates subtypes of a problem and how each problem subtype would have different semantic slots to apply math rules. The concept of \textit{part-whole} has two slots, one for the whole that accepts a single variable and the other for its parts that accepts a set of variables of size at least two. The \textit{change} concept has four slots, namely start, end, gains, losses which respectively denote the original value of a variable, the final value of that variable, and the set of increments and decrements that happen to the original value of the variable. The \textit{comparison} concept has three slots, namely the large quantity, the small quantity, and their difference  \cite{mitra2016learning}. Similarly, other such mutually exclusive groups were identified by researchers to categorize MWPs, like  (i) join-separate, (ii) part-whole and (iii) compare \cite{amnueypornsakul2014machine}; (i) change, (ii) combine (iii) compare \cite{bakman2007robust}; or (i) transfer, (ii) dimensional analysis, (iii) part-whole relation, (iv) explicit math, \cite{roy2018mapping} etc.

Commonsense and linguistic knowledge could be useful to derive the semantic structure and represent the meaning of MWPs. \cite{briars1984integrated} investigates how commonsense knowledge could be helpful to solve complex problems narrating real-world situations. The knowledge that an object defined in the MWP is a member of both a set and its superset and whether subsets can be exchanged in the context of deriving an answer for an MWP. For example, the math word problem shown in figure \ref{fig:mwp} illustrates objects “orange” and “apple.” Both belong to a superset “fruits,” and they can be exchanged if the unknown quantity in the question is agnostic to the type of fruit. Dolphin language representation of MWPs requires extracting nodes and their relationships, where nodes could be constants, classes, or functions. Commonsense knowledge is used to classify entities sharing common semantic properties \cite{shi2015automatically}. Linguistic knowledge like verb sense and verb entailment could help efficiently semantic parsing of MWPs narratives to extract semantic slots and fill their values. Researchers have explored such external knowledge from WordNet \cite{hosseini2014learning} or E-HowNet \cite{chen2005extended} \cite{lin2015designing}.

\section{Neural Approaches}
The recent advancement in deep learning approaches has opened up new possibilities. There is a flurry of research that aims to apply neural networks to solve MWPs. Sequence to sequence architectures, transformers, graph neural networks, convolutions, and attention mechanisms are a few such architectures and techniques of neural approaches. To empower neural approaches, larger and diverse datasets have been also created \cite{amini2019mathqa}, \cite{wang2017deep}, \cite{zhang2020graph}, \cite{saxton2019analysing}, \cite{lample2019deep}. Also, the potential of neural approaches solving complex problems on calculus and integration has raised expectations on its future promise. These approaches could be further categorized based on their problem formulation, (i) to predict answers directly, (ii) generate intermediate math expressions, or (iii) retrieve a template. However, it is evident that neural networks are black boxes, and hard to interpret their functioning and explain their decision \cite{gaur2021semantics}. Attempts to interpret its functioning reveal how volatile its reasoning ability is, and they lack generalizability \cite{patel2021nlp}.

In the following sections, we aim to categorize the current state-of-the-art development. Then, we navigate problem formulations, datasets and analysis, approaches and architectural design choices, data augmentation methods, and interpretability and reasoning ability.

\begin{table*}[hbt!]
\fontsize{9pt}{13.5pt}
\selectfont
\begin{tabular}{p{0.25\linewidth} | p{0.23\linewidth} |  p{0.07\linewidth} | p{0.06\linewidth}| p{0.18\linewidth} | p{0.06\linewidth}}
\hline Dataset & Mathematics Area & Language & Type & Annotation & Size \\
\hline
AI2 \cite{hosseini2014learning} & Arithmetic & English & Curated & Equation / Answer & 395 \\
IL \cite{roy2015reasoning} & Arithmetic & English & Curated & Equation / Answer &	562	\\
ALGES \cite{koncel2015parsing} & Arithmetic	& English & Curated & Equation / Answer & 508 \\
AllArith \cite{roy2017unit} & Arithmetic & English & Derived & Equation / Answer	& 831 \\
\hline
Alg514 \cite{kushman2014learning} & Algebraic (linear) & English & Curated &	Equation/Answer & 514 \\			
Dolphin1878 \cite{shi2015automatically} & Algebraic (linear) & English & Curated &	Equation / Answer & 1,878 \\			
DRAW \cite{upadhyay2015draw} & Algebraic (linear) & English & Curated & Equation / Answer / Template & 1,000 \\
Dolphin18K \cite{huang2016well} & Algebraic (linear, nonlinear) & English & Curated &	Equation / Answer & 18,000	\\
\hline
MAWPS \cite{koncel2016mawps} & Arithmetic, Algebraic & English & Curated & Equation/answer & 3,320 \\	
AQuA \cite{ling2017program} & Arithmetic, Algebraic (linear, nonlinear) & English & Curated & Rationale/MCQ-choices/Answer & 100,000 \\	
MathQA \cite{amini2019mathqa} & Arithmetic, Algebraic & English & Derived &	Equation/MCQ-choices/Answer	& 37,000 \\
MATH \cite{hendrycks2021measuring} & Algebra,
Number Theory, Probability, Geometry, Calculus
 & English & Curated & Step-by-step solution / Answer & 12500 \\
 \hline
ASDiv \cite{miao2021diverse} & Arithmetic, Algebraic & English & Derived &	Equation/Answer +
 Grade/Problem-type &	2,305 \\
SVAMP \cite{patel2021nlp} &	Arithmetic & English & Derived &	Equation/Answer	& 1,000 \\
\hline
Math23K	\cite{wang2017deep} & Algebraic (linear) & 
Chinese & Curated & Equation/Answer	& 23,161	\\
HMWP \cite{qin2020semantically}	& Algebraic (linear + nonlinear) & Chinese & Curated & Equation/Answer & 5,491 \\
Ape210K	\cite{zhao2020ape210k} & Algebraic (linear)	& Chinese & Curated & Equation/Answer & 210,488 \\	
\hline
Deepmind Mathematics \cite{saxton2019analysing} & Algebra, Probability, Calculus & English & Synthetic & Answer & 2,000,000	\\
Integration \& Differentiation Synthetic Dataset \cite{lample2019deep} & Integration, Differentiation & English & Synthetic	& Answer & 160,000,000 \\
AMPS \cite{hendrycks2021measuring} & Algebra, Calculus, Geometry, Statistics, Number Theory & English & Synthetic & Step-by-step solution / Answer & 5,000,0000 \\
\hline
\end{tabular}
\caption{Categorization of MWP Datasets available for  training AI models based on (a) Mathematics Area, (b) Language, (c) Type, and (d) Annotation.}
\label{tab:MWPDatasets}
\end{table*}

\subsection{Problem Formulation}

We can categorize the problem formulation to solve MWPs in three different ways, (i) predicting the answer directly, (ii) generating a set of equations or mathematical expressions, and inferring answers from them by executing them, and (iii) retrieving the templates from a pool of templates derived from training data and augmenting numerical quantities to compute the answer. An example is shown in Figure \ref{fig:mwp}.

The first approach to predict the answers directly could demonstrate the inherent ability of the neural model to learn complex mathematical transformations \cite{csaji2001approximation}; however, the black-box nature of such models suffers from poor interpretability and explainability \cite{gaur2021semantics}.   It has been shown that sequence to sequence (seq-to-seq) neural models could compute free form answers for complex problems with high accuracy when trained on large datasets \cite{saxton2019analysing}\cite{lample2019deep}.  It has been analyzed how models behave when just numerical parameters are changed vs. mathematical concepts in the test set. \cite{lample2019deep} achieved near-perfect accuracy of 99\% mainly attributed to the learning ability of neural models over a huge dataset. \cite{ran2019numnet} proposes a numerically aware graph neural network to directly predict the type of answer and the actual answer.

The second approach deals with generating an expression tree and executing them to compute the answer. Such models have relatively better interpretability and explainability and generated expressions, which provides scaffolding to the reasoning ability of the model. Seq-to-seq neural models framing from LSTM, Transformers, GNN and Tree decoders have proven to useful \cite{wang2017deep} \cite{amini2019mathqa} \cite{xie2019goal} \cite{qin2020semantically} \cite{liang2021mwp}. The key challenge for methods in this approach is the need for expert annotated datasets, as they need expression trees as the labels for each problem in addition to its answer \cite{amini2019mathqa}\cite{wang2017deep}\cite{koncel2016mawps}.

The third approach derives template equations from training data, retrieves the most similar template, and substitutes numerical parameters. This approach suffers from generalization ability, as the set of templates are limited to the training set. Such methods would fail to solve diverse and out-of-corpus MWPs as they would not be able to retrieve the template needed to solve them. It was one of the popular statistical machine learning approach \cite{kushman2014learning} \cite{koncel2015parsing}, \cite{mitra2016learning}, but there are also few other neural approaches, which have studied its value for an ensemble setup or standalone model with larger training data \cite{wang2017deep} \cite{robaidek2018data}. 

Generally, these neural approaches follow encoder-decoder architecture to predict expression trees or directly compute the answer, specifically when the answer could be an expression in itself. We further explain different variants of encoder-decoder architectures in section 4.3.

\subsection{Analyzing Datasets}

One of the current limitations of achieving performance similar to human-level mathematical intelligence is the availability of labeled datasets. Table \ref{tab:MWPDatasets} categorizes various MWPs datasets along with information like dataset size, language, type of dataset, mathematics area etc. Each dataset focus on specific areas of Mathematics, i.e., \textit{Arithmetic, Algebra, Calculus, etc}. AI2, IL, ALGES, ALLArith, SVAMP are examples of datasets for arithmetic MWPs. Datasets such as Alg514, Dolphin1878, DRAW, Dolphin18K, Math23K, HMWP, and Ape210K focus on Algebraic problems. MAWPS, AQuA, MathQA, and ASDiv are examples of datasets covering MWPs from \textit{Arithmetic and Algebra}. There have been efforts to create MWPs datasets for complex areas including \textit{Probability, Integration and Differentiation} \cite{saxton2019analysing} \cite{lample2019deep}.  \ref{tab:MWPDatasets} covers examples of MWP datasets in English and Chinese languages. 

MWPs datasets are either curated from educational resources, derived by processing other available datasets or synthetically created. For example., Alg514 is created by curating problems from algerba.com. Similarly, Dolphin1878 comprises of curated problems from both \textit{algebra.com} and \textit{answers.yahoo.com}. Derived datasets have been derived by processing existing datasets to address concerns related to training corpus size \cite{roy2017unit}, equation annotation \cite{amini2019mathqa},  lexical diversity \cite{miao2021diverse}, problem types \cite{roy2017unit} or adversarial reasoning ability \cite{patel2021nlp}. Synthetic datasets could be very large, but they lack generalization. They are ultimately driven from deterministic sets of templates or rules and combinations of them for specific math areas \cite{saxton2019analysing} \cite{lample2019deep} \cite{hendrycks2021measuring}. Synthetic datasets could also be used in pre-training neural models. Curated datasets are generally small in size, e.g., AI2, IL, Alg514, DRAW, MAWPS, etc., or suffer from inconsistencies, e.g., AQuA or MathQA. Neural models generating a set of equations or expression trees would need a dataset, where MWPs are labeled with a set of equations or expression trees. Many of these curated datasets would not have associated annotated expression trees; hence these datasets would require manual effort to annotate. If datasets are created by curating relevant webpages, it would also possibly need answer extraction from free-form explanations \cite{huang2016well}. Equations for MWPs could also be extracted from their solutions on webpages using pattern matching, but which could lead to some inconsistencies \cite{amini2019mathqa}.  On the other hand, it is very costly to augment these datasets with expert annotations. There can be multiple possible ways to write a semantically unique equation, which demands equation normalization to make sure labels are normalized \cite{huang2018neural}.  

There is an active interest among the research community to analyze and improve the quality of datasets. Datasets like AsDiv and SVAMP are derived to address the quality concerns by processing existing datasets. Lexically similar questions would lead the model to learn the heuristic patterns to predict the answer \cite{miao2021diverse}. Lexicon Diversity Score measures the lexical diversity of questions provided in the dataset, and datasets with better lexicon diversity score would be more reliable to test a model's performance \cite{miao2021diverse}. Also, adversarial approaches further extend such datasets with adversarial variants to test the mathematical reasoning ability \cite{patel2021nlp}. 

\subsection{Encoder Decoder Architectures}

\begin{figure*}[h]
    \centering
    \includegraphics[scale=0.5]{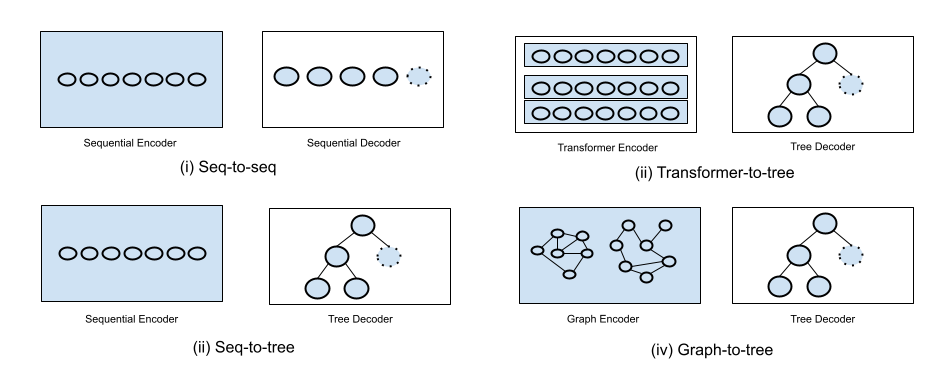}
    \caption{Popular Neural Architectures for Solving Math Word Problems}
    \label{fig:neural_arch}
\end{figure*}

\cite{wang2017deep} proposed the first encoder-decoder architecture to generate equations or expression trees for MWPs. It used GRU based encoder and LSTM based decoder. Since then, researchers have actively sought to explore the optimal architecture to get better performance for predicting equations or expression trees on diverse datasets comprising complex MWPs. As datasets are limited in size, researchers kept iterating to find better architectures to learn semantic structures of math word problems well. These novel neural architectures can be grouped into four categories, \textbf{(1) seq-to-seq, (2) seq-to-tree, (3) transformer-to-tree, (4) graph-to-tree}.
 
Sequence-to-sequence architecture is the default choice of any natural language generative model to transform a natural language input to output. Output sequence is either free form answer expression, equations, or traversal sequence of expression trees \cite{griffith2021solving} \cite{amini2019mathqa}. \cite{wang2017deep} leverages GRU-based encoder and LSTM based decoder, \cite{amini2019mathqa} uses attention mechanisms over RNNs, and conditions decoder with category of MWPs. For larger synthetic datasets \cite{lample2019deep} \cite{saxton2019analysing}, there isn't a need for semantically rich hybrid architectures as seq-to-seq models could achieve good performance. Transformers outperforms attentional LSTMs on larger datasets \cite{saxton2019analysing}, mainly attributed to more computations with the same numbers of parameters, internal memory of transformers which is pre-disposed to mathematical objects like quantities, and better gradient propagation with shallower architectures. There has been an effort to model situations narrated in MWPs using agents, attributes, and relations and train a seq-to-seq model to convert relation text to equations \cite{hong2021smart}.

Sequence-to-tree, transformer-to-tree, and Graph-to-tree architectures attempt to incorporate the tree structure of the output mathematical equations or expressions into it. The decoder in such models is a tree decoder, where it generates the next token by taking hidden states of its parent and sibling and its subtree (if any) into the account \cite{xie2019goal}\cite{liu2019tree}\cite{qin2020semantically}\cite{liang2021mwp} \cite{wu2021math}. Attention-based architectures leverage attention mechanisms to attend only parents and siblings \cite{qin2020semantically}\cite{liang2021mwp}. Operators would be parent nodes, and operands would be their children nodes, which would recursively expand further. Encoders in such architectures varied from GRU/LSTM based \cite{xie2019goal} or transformer-based \cite{liang2021mwp}. As tree decoders help model learning the tree semantics out output expressions, such models perform better than the seq-to-seq model on datasets with limited size. The expected output could be represented as a computational tree, which is then traversed and evaluated to compute the answer.

Graph-to-tree architecture further exploits the graph semantics present in MWPs \cite{ran2019numnet} \cite{zhang2020graph} \cite{li2019modeling}. Graph semantics captures the relationships between different entities, which can be thought of as relationships among numerical quantities, their description, and unknown quantities. The goal-driven tree-structured model is designed to generate expressions using computational goals inferred from an MWP, which uses graph transformer network as the encoder to incorporate quantity comparison graph and quantity cell graph \cite{zhang2020graph}. It leverages self-attention blocks for incorporating these instances of graphs. The numerical reasoning module attempts to learn comparative relations between numerical quantities by connecting them in a graph. Encoder representations are appended with representations learned by numerical reasoning module to improve the performance \cite{ran2019numnet}. An MWP can be segmented into a quantity spans and a question span. Specific self-attention head blocks which could just attend quantity span, question span, or both could be used to learn graph relationships between quantities and question  present in an MWP \cite{li2019modeling}.

\subsection{Design Choices}

In this section, we aim to highlight several design decisions for the above architectures. First, the seq-to-seq model suffers from generating spurious numbers or predicting numbers at the wrong position. Copy and alignment mechanisms can be used to avoid this problem \cite{huang2018neural}. Second, it has become a common practice to keep decoder vocabulary limited to numbers, constants, operators, and other helpful information present in the question \cite{xie2019goal} \cite{zhang2020graph}.

There are multiple ways of generating output expressions, and maximum likelihood expectations would compromise learning if predicted, and expected expressions are different but semantically the same. Reinforcement learning would solve this problem by rewarding the learning strategy based on the final answer \cite{huang2018neural}.

Tree-based decoders have received lots of attention because of their ability to invoke tree relationships present in mathematical expressions. Tree decoders attend both parents and siblings to generate the next token. Bottom-up representation of subtree of a sibling could further help to derive better outcomes  \cite{qin2020semantically}. For efficient implementation of tree decoder, the stack data structure can be used to store and retrieve hidden representations of parent and sibling  \cite{liu2019tree}. Tree regularization transforms encoder representation and subtree representations and minimizes their L2 distance as a regularization term of the loss function \cite{qin2020semantically}.

Graph-based encoder aims to learn different types of relationships among the constituents of MWPs. There are several attempts to construct different types of graphs. \cite{ran2019numnet} inserts a numerical reasoning module that uses numerically aware graph neural networks, where two types of edge less-than-equal-to and greater-than connect numerical quantities present in the question. A situation model for algebra story problem SMART aims to build a graph using attributed grammar, which connects nodes with its attributes using relationships extracted from the problem \cite{hong2021smart}. Self-attention blocks of transformers could also be used to model the graph relationships \cite{zhang2020graph} \cite{li2019modeling}. The goal-driven tree-structured model incorporated quantity cell graph and quantity comparison graph \cite{zhang2020graph}. Quantity cell graph connects numerical quantity of the input equations with its descriptive words or entities, whereas quantity comparison graph is similar to the concept used by  \cite{ran2019numnet}. Segmentation of question into quantity span and question span, and using self-attention blocks to derive global attention, quantity-related attention, quantity pair attention, and question-related attention is another such approach to derive better semantic representation \cite{li2019modeling}.

Multitasking is a prevalent paradigm to train the same model for multiple tasks. It enriches the semantic representations of models and avoids them getting overfitted. Auxiliary tasks could also be part of such a setup. Auxiliary tasks like Common Sense Prediction Task, Number Quantity Prediction, and Number location prediction could be helpful \cite{qin2021neural}. The Commonsense prediction task aims to predict the commonsense required to solve an MWP, like the number of legs of a horse, or a number of days in a week, etc. On the other hand, mask language modeling pre-training on mathematical corpus Ape210k in a self-supervised way to help its downstream application for math word problems \cite{liang2021mwp}.

\subsection{Data Augmentation and Weak Supervision}

Large datasets have empowered neural models to learn complex mathematical concepts like integration, differentiation \cite{lample2019deep} \cite{saxton2019analysing}, etc, but such models often overfit on smaller datasets \cite{miao2021diverse} \cite{patel2021nlp}. Data augmentation is a popular preprocessing technique to increase the size of training data. Generally, data augmentation creates the variant of training records by applying domain-specific argumentation techniques. Several techniques were proposed for the data augmentation of MWP datasets. Reverse operation-based augmentation techniques swap an unknown quantity with a known quantity in an MWP, and their expression trees are modified to reflect the change \cite{liu2020reverse}. Different traversal orders of expression trees and other preprocessing techniques like lemmatization, POS tagging, sentence reordering, stop word removal, etc., and their impact on the performance of models have been studied by \cite{griffith2021solving}. Data processing and augmentation could also help to generate adversarial datasets, which could test the reasoning ability of models \cite{patel2021nlp}\cite{miao2021diverse}.

Weak supervision is another popular technique that generates large datasets with noisy labels. The model training process could cancel noise and learn actual patterns. For example, generating expression trees for an MWP dataset with question and answer could be done using weak supervision, where expression trees may not always be accurate. Still, the answer it evaluates would match the desired answer.  Learning by fixing is a technique where expression trees are generated iteratively by fixing operators and operands till it evaluates to the desired answer \cite{hong2021learning}. Such a technique helps model learning the more diverse ways of solving mathematical problems.

\section{Mathematical Reasoning in Neural Approaches}

Mathematical reasoning is core to human intelligence. As explained before, it is a complex phenomenon interfacing natural language understanding and visual understanding to invoke mathematical transformations. Over the past few decades, there has been a flurry of research to develop models with mathematical reasoning ability; however, the progress is limited. Larger datasets definitely help models learning to solve complex and niche problems \cite{lample2019deep} \cite{saxton2019analysing}, but a common purpose model to solve a diverse variety of math problems is still a distinct reality \cite{patel2021nlp} \cite{miao2021diverse}. Smaller datasets overfit the models, and hence reasoning ability is questionable. Non-neural models could not improve performance with larger datasets \cite{huang2016well}.
On the other hand, neural models achieve much better performance on the smaller datasets and attain near-perfect performance over very large datasets; however, its BlackBox structure hinders interpreting its reasoning ability. There is a decent consensus among researchers to solve problems by first generating intermediate expression trees and evaluating them using solvers to predict the answer \cite{wang2017deep} \cite{amini2019mathqa} \cite{xie2019goal}. As opposed to directly predicting the final answer, generating intermediate expression trees helps understand models' reasoning ability better.  Currently, the efforts are mainly to solve MWPs of primary or secondary schools as shown in Table \ref{tab:MWPDatasets}. The state of the current art is still very far from building mathematical intelligence that could also infer visual narratives and natural language and use domain-specific knowledge.

Models with desired reasoning ability would not only be generalizable, but it would also help to make systems interpretable and explainable. On the other hand, interpretable and explainable systems are easier to comprehend, which helps to progress further by expanding the scope of its reasoning on more complex mathematical concepts. The field of mathematics is full of axioms, rules, and formulas, and it would also be important to plug explicit and definite knowledge into such systems. 

\subsection{Interpretability, and Explainability}

Deep learning models directly predicting answers from an input problem lack interpretability and explainability. As models often end up learning shallow heuristic patterns to arrive at the answer. \cite{huang2016well}, \cite{patel2021nlp}. That’s where it is important to generate intermediate representations or explanations, then infer the answer using them. It not only improves interpretability but also provides scaffolding to streamline reasoning ability and hence generalizability. Such intermediate representations could be of different forms, and they could be interspersed with natural language explanations to aid the explainability of the model further. 

\subsubsection{Intermediate Representations}
The semantic parsing approach attempts to derive semantic structure from the problem narratives. Still, they are not generalizable because of the inherent limitations of statistical and classical machine learning techniques to represent the meaning. However, deriving intermediate representations and inferring the solution henceforth would remain one of the niche areas to explore in the context of neural approaches. Transforming problem narratives to expressions or equations and then evaluating them to get an answer is one of the most important approaches suitable for elementary level MWPs \cite{wang2017deep}. There were other attempts in deriving representation language \cite{shi2015automatically}, using logic forms \cite{liang2018meaning} and intermediate meaning representations \cite{huang2018using}, etc. It would require much more effort to derive intermediate representation for comprehensive and complex mathematical forms and expressions in hindsight.

\subsubsection{Interspersed Natural language Explanations}

Natural language explanations to describe the mathematical rationale would reflect the system's reasoning ability and also makes the system easier to comprehend and explainable. For example, \cite{ling2017program} uses LSTM to focus on generating answer rationale, which is a natural language explanation interspersed with algebraic expression to arrive at a decision. AQuA dataset contains 100,000 such algebraic problems with answer rationales. Such answer rationales not only improve interpretability but provide scaffolding which helps the model learn mathematical reasoning better. Language models like WT5, which produce explanations along with prediction, could be an inspiration \cite{narang2020wt5}. It would be essential to extend such approaches for more complex mathematical concepts beyond algebra. Natural language explanations interspersed with mathematical expressions would be critical areas for the researchers to focus on for building more comprehensive systems. Such interspersed explanations would open up the possibility to help students understand how to solve the problem more effectively or would foster users' trust in the system.

\subsection{Infusing Explicit and Definitive Knowledge}

Mathematical problem solving requires identifying relevant external knowledge and applying it to the mathematical rules to arrive at an answer. External knowledge could be commonsense world knowledge, particular domain knowledge, or complex mathematical formulas. There is some progress to infuse commonsense world knowledge \cite{liang2021mwp} or predict it using auxiliary tasks \cite{qin2021neural}, however remaining two dimensions are yet to be addressed. Such external knowledge is incorporated using rule-based, pattern-matching approaches in non-neural models; on the other hand, neural models leverage data augmentation and auxiliary task training. \cite{zhang2020graph} \cite{xie2019goal} developed novel neural architectures to represent semantics captured in the problems using graph and tree relations. These examples are of the shallow knowledge infusion category. It would be intriguing to explore deep knowledge infusion wherein model architecture would transform and leverage external knowledge vector spaces during its training \cite{wang2020k} \cite{faldu2021ki}. Furthermore, it would be promising to build an evaluation benchmark for such knowledge-intensive mathematical problems to encourage and streamline efforts to solve math problems \cite{sheth2021knowledge}.

\subsection{Reinforcement Learning}

Mathematical reasoning ability comprises applying a series of mathematical transformations. Rather than learning mathematical transformations, the system could focus on synthesizing the computational graph. Learning algorithms like gradient descent works on iterative reduction of the error. The choices of these mathematical transformations are not differential, and hence it is unclear if its the optimal strategy. Learning policy to construct such a computation graph using reinforcement learning could be helpful \cite{palermo2021reinforcement} \cite{huang2018neural}. Exponential search space of mathematical concepts is another key problem where reinforcement learning could be useful \cite{wang2018mathdqn}. Reinforcement learning reduces a problem to state transition problem with reward function. It initializes the state from a given problem, and based on the semantic information extracted from the problem, it derives actions from updating the state with a reward or penalty. It learns a policy to maximize the reward as it traverses through several intermediate states. For an MWP, the computation graph of expression trees denotes the state, and action denotes the choices of mathematical transformations and their mapping with operands from the input. The evaluation of the computation graph would give an answer, and a penalty or reward would be given based on the comparison between the computed answer and the expected ground truth. Researchers have also leveraged reinforcement learning on difficult math domains like automated theorem proving \cite{crouse2021deep} \cite{kaliszyk2018reinforcement}. It is still an early stage of the application of reinforcement learning for solving math word problems. It could bring the next set of opportunities to build systems capable of mathematical reasoning ability. 

\section{Outlook}

The mathematical reasoning process involves understanding natural language and visual narratives and extracting entities with explicit and implicit quantities building a self-consistent symbolic representation invoking mathematical laws, axioms, and symbolic rules.

The ability to solve MWPs in an automated fashion could lead to many applications in the education domain for content creation and delivering learning outcomes \cite{donda2020framework} \cite{faldu2020adaptive}.

We surveyed sets of different approaches and their claims to have reasonable solved MWPs on specific datasets. However, we have also highlighted studies with “concrete evidence” that existing MWP solvers tend to rely on shallow heuristics to achieve their high performance and questions these models’ capabilities to solve even the simplest of MWPs robustly \cite{huang2016well} \cite{patel2021nlp} \cite{miao2021diverse}.

Non-neural approaches do not improve linearly with larger training data \cite{huang2016well}, on the other hand, neural approaches have shown a promise to solve even complex MWPs trained with very large corpus \cite{lample2019deep} \cite{saxton2019analysing}. However, the availability of extensive training data with diverse MWPs is a crucial challenge. The inherent limitation of neural models in terms of interpretability and explainability could be addressed partly by predicting expression trees instead of directly predicting answers \cite{gaur2021semantics}. Such expression trees could be converted into equations, and which are evaluated to infer the final answer. Further, expression trees also provide scaffolding to assist mathematical reasoning ability and open up design choices like graph-based encoder and tree-based decoder, which helps to learn from the inadequately sized training corpus. Interspersed natural language explanations and infusing explicit knowledge could be of significant value as it not only improves the explainability of the system but also guides the model to traverse through intermediate states anchoring mathematical reasoning \cite{ling2017program} \cite{gaur2021semantics}. Interspersed explanations not only solve the math word problems but also engage users on how to solve them. 

We highlight that solving MWPs would require the inherent ability of mathematical reasoning, which comprises natural language understanding, image understanding, and the ability to invoke domain knowledge of mathematical laws, axioms, and theorems. There is a massive scope of complementing neural approaches with external expertise/knowledge and developing design choices as we extend the problem to complex mathematical areas.  

\bibliography{mwpsurvey}
\bibliographystyle{acl_natbib}

\end{document}